\documentclass{article}
\usepackage{spconf,amsmath,graphicx}
\usepackage{amssymb}
\usepackage{amsmath}
\usepackage{mathtools}
\usepackage{xcolor}
\usepackage{multirow}
\usepackage[ruled,vlined]{algorithm2e}

\usepackage{kotex}
\usepackage{color}

\title{Mutually-Constrained Monotonic Multihead Attention for Online ASR}
\name{Jaeyun Song\qquad Hajin Shim\qquad Eunho Yang 
\thanks{
Copyright 2021 IEEE. Published in ICASSP 2021 - 2021 IEEE International Conference on Acoustics, Speech and Signal Processing (ICASSP), scheduled for 6-11 June 2021 in Toronto, Ontario, Canada. Personal use of this material is permitted. However, permission to reprint/republish this material for advertising or promotional purposes or for creating new collective works for resale or redistribution to servers or lists, or to reuse any copyrighted component of this work in other works, must be obtained from the IEEE. Contact: Manager, Copyrights and Permissions / IEEE Service Center / 445 Hoes Lane / P.O. Box 1331 / Piscataway, NJ 08855-1331, USA. Telephone: + Intl. 908-562-3966.
}}
\address{KAIST, Daejeon, South Korea}
\begin{document}
%
\maketitle
\begin{abstract}
Despite the feature of real-time decoding, Monotonic Multihead Attention (MMA) shows comparable performance to the state-of-the-art offline methods in machine translation and automatic speech recognition (ASR) tasks.
However, the latency of MMA is still a major issue in ASR and should be combined with a technique that can reduce the test latency at inference time, such as head-synchronous beam search decoding, which forces all non-activated heads to activate after a small fixed delay from the first head activation. In this paper, we remove the discrepancy between training and test phases by considering, in the training of MMA, the interactions across multiple heads that will occur in the test time. Specifically, we derive the expected alignments from monotonic attention by considering the boundaries of other heads and reflect them in the learning process. We validate our proposed method on the two standard benchmark datasets for ASR and show that our approach, MMA with the mutually-constrained heads from the training stage, provides better performance than baselines.  


\end{abstract}
\section{Introduction}
\label{sec:introduction}

Online automatic speech recognition (ASR), which immediately recognizes incomplete speeches as humans do, is emerging as a core element of diverse ASR-based services such as teleconferences, AI secretaries, or AI booking services. In particular, in these days, where the untact service market is rapidly growing due to the recent global outbreak of COVID-19, the importance of providing more realistic services by reducing latency is also growing. However, of course, online ASR models \cite{graves2006connectionist,graves2012sequence} targeting real-time inference have concerns about performance degradation compared to traditional DNN-HMM hybrid models with pre-segmented alignments or offline models based on Transformer \cite{DBLP:conf/nips/VaswaniSPUJGKP17,DBLP:conf/icassp/DongXX18}, which is the state-of-the-art in many sequence-sequence tasks nowadays.

In order to overcome this performance-delay trade-off, several attempts have been made to learn or find monotonic alignments between source and target via attention mechanism \cite{DBLP:conf/nips/ChorowskiBSCB15,DBLP:conf/nips/JaitlyLVSSB16,DBLP:conf/icassp/MoritzHR19,DBLP:conf/icml/RaffelLLWE17, DBLP:conf/iclr/ChiuR18}. Especially, Monotonic Attention (MA), and Monotonic Chunkwise Attention (MoChA) \cite{DBLP:conf/icml/RaffelLLWE17, DBLP:conf/iclr/ChiuR18} learn alignments in an end-to-end manner by calculating differentiable expected alignments in training phase and shows comparable performance to models using an offline attention.

Very recently, motivated by the success of Transformer architecture even in ASR \cite{DBLP:conf/icassp/DongXX18}, direct attempts to make it online by applying these learning alignment strategies to Transformer, not on the traditional RNN based models, are emerging \cite{DBLP:conf/icassp/MoritzHR20,DBLP:conf/icassp/Dong020,tsunoo2020streaming,DBLP:conf/icassp/MiaoCGZ020}. Among others, Monotonic Multihead Attention (MMA) \cite{DBLP:conf/iclr/MaPCPG20} converts each of multi-heads in Transformer to MA and exploits the diversity of alignments from multiple heads. In order to resolve the issue of MMA that has to wait for all multi-heads to decode, HeadDrop \cite{inaguma2020enhancing} drops heads stochastically in the training stage. \cite{inaguma2020enhancing} also proposed to use head-synchronous beam search decoding (HSD) which limits the difference in selection time  between the heads in the same layer only in the inference phase, but resulting in the discrepancy between training and inference.
In this paper, we propose an algorithm, called ``Mutually-Constrained Monotonic Multihead Attention" (MCMMA), that enables the model to learns alignments along with other heads by modifying expected alignments to consistently bring constrained alignments of the test time to the training time. 
By bridging the gap between the training and the test stages, MCMMA effectively improves performance. 
\vspace{-.3cm}
\section{Preliminary}
\label{sec:preliminary}

We first review the main components which our model is based on, including monotonic attention, monotonic multihead attention, and HeadDrop with head-synchronous beam search decoding in Subsection \ref{ssec:ma}, \ref{ssec:mma}, and \ref{ssec:hsd}, respectively.
\vspace{-.3cm}
\subsection{Monotonic Attention}
\label{ssec:ma}

MA \cite{DBLP:conf/icml/RaffelLLWE17} is the attention-based encoder-decoder RNN model which is able to learn monotonic alignments in an end-to-end manner. The encoder processes input sequence $\mathbf{x}=(x_1,\ldots,x_T)$ to encoder states $\mathbf{h}=(h_1,\ldots,h_T)$. At $i$-th output step, the decoder sequentially inspects encoder states from the last selected one in the previous step and decide whether to take it or not to produce current output. The probability $p_{i, j}$ to select $h_j$ for $i$-th output is computed as
\begin{align*}
    &e_{i,j}=\text{MonotonicEnergy}\left(s_{i-1},h_{j}\right) \text{ and } p_{i,j}=\sigma\left(e_{i,j}\right)
\end{align*}
where $s_{i-1}$ is a decoder state of $(i-1)$-th output step.
If $h_j$ is selected, the RNN decoder takes it as context $c_i = h_j$ with the previous decoder state $s_{i-1}$ and output $y_{i-1}$ to compute current state.

To make alignment learnable in the training phase, a hard selected context above is replaced by a expected context $c_i = \sum_{j=1}^{L}{\alpha}_{i,j}h_j$, the weighted sum of $\mathbf{h}$ with the expected alignment $\alpha$
computed as 
\begin{align}\label{eq:alpha}
    {\alpha}_{i,j}=p_{i,j}\sum_{k=1}^{j}\left({\alpha}_{i-1,k}\prod_{l=k}^{j-1}\left(1-p_{i,l}\right)\right). 
\end{align}

MoChA \cite{DBLP:conf/iclr/ChiuR18} extends MA by performing soft attention over fixed-length chunks of encoder states preceding the position chosen by a MA mechanism.

\subsection{Monotonic Multihead Attention}
\label{ssec:mma}

MMA \cite{DBLP:conf/iclr/MaPCPG20} applies MA mechanism to Transformer \cite{DBLP:conf/nips/VaswaniSPUJGKP17} by making each of the multiple heads of decoder-encoder attention learn monotonic alignments as MA.
MMA borrows scaled dot-product operation of Transformer. 

Although MMA leads to considerable improvement in online machine translation, the latency is still high since the model should wait until all heads to select their contexts for every decoding step.

Thus, the authors of \cite{DBLP:conf/iclr/MaPCPG20} proposed to use additional regularization to minimize the variance of expected alignments of all heads to reduce the latency. 
Nevertheless, this approach does not model the dependency between heads explicitly. 


\subsection{HeadDrop and Head-Synchronous Decoding}
\label{ssec:hsd}

HeadDrop \cite{inaguma2020enhancing} is the method that drops each head stochastically for each individual to learn alignments correctly. 
This approach improves boundary coverage and streamability of MMA \cite{inaguma2020enhancing}.
Head-synchronous decoding (HSD) \cite{inaguma2020enhancing} is the inference algorithm, where the leftmost head forces slow heads, which fail to choose any frames within waiting time threshold $\epsilon$, to choose the rightmost of selected frames.
However, HSD considers alignments of other heads and only at the test phase. 

\begin{figure}
\begin{center}
\includegraphics[width=1\linewidth]{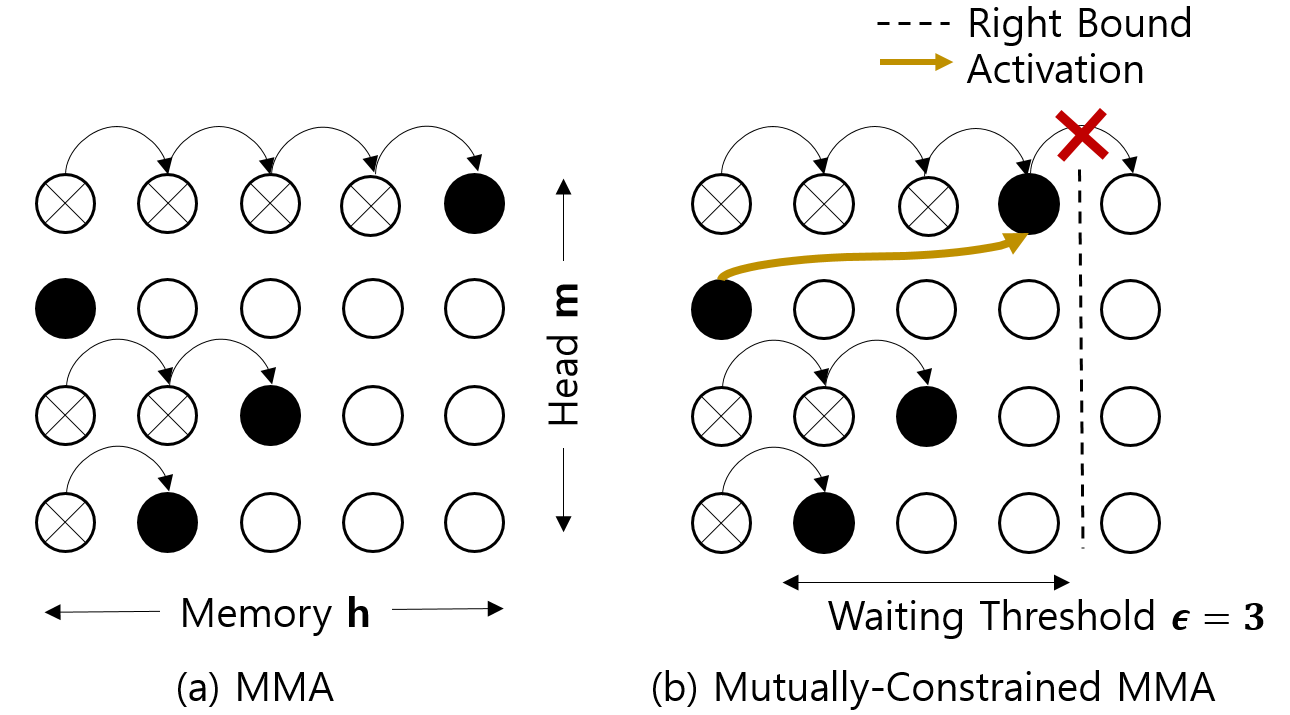}
\end{center}
\caption{(a) represents the monotonic attention in MMA \cite{DBLP:conf/iclr/MaPCPG20}. MMA does not consider alignments of other heads. (b) is our approach considering alignments of other heads. The selection of the first head is forced by the second head. The inference algorithm is the same with HSD \cite{inaguma2020enhancing}.}
\label{fig:concept}
\end{figure}

\section{Method}
\label{sec:method}

In this section, we propose the algorithm to learn alignments of MA heads under a constraint of the difference of heads' alignments within each decoder layer.
To reduce latency, our approach forces MMA heads 
to select input frames within a limited range, not only in the testing phase but also in the training phase by estimating differentiable expected alignments of MA heads under constraints. With newly estimated expected alignments, we follow the overall training and test process of \cite{inaguma2020enhancing}.
Before we address the details, we define two functions for the convenience of deriving the expectations of constrained alignments.

Let $\epsilon$ be the waiting threshold for the training stage, and ${\alpha}^m_{i,T+1}$ be the probability for the $m$-th head not to select any frames where $T$ is the length of the input sequence. 
We define function $A^m_{i,j}$ which is equal to $\alpha_{i, j}$ when it takes expected alignment $\alpha$ as an argument (See Eq. (\ref{eq:alpha})):
\begin{alignat}{1}
    &{A}^m_{i,j}\left(\alpha\right)=p^m_{i,j}\sum_{k=1}^{j}\left({\alpha}^m_{i-1,k}\prod_{o=k}^{j-1}\left(1-p^m_{i,o}\right)\right). \label{eq:newalpha}
\end{alignat}
The probability $B^m_{i, j}$ that the $m$-th head does not choose any frames until reaching $h_j$ for $i$-th output can be represented as a function taking $\alpha$:
\begin{alignat}{1}
    &{B}^m_{i,j}\left(\alpha\right) = \left(  1-\sum^j_{k=1}{\alpha}^m_{i,k} \right).
\end{alignat}


Before we consider interdependent alignments of all heads under the constraints, we first consider the simple situation that each head chooses a frame within a waiting time threshold $\epsilon$ from the last selected frame at the previous step. 
We define another function $C$ for expected alignments as
\begin{alignat}{1}
    &{C}^m_{i,j}\left(\gamma\right) = {A}^m_{i,j}\left(\gamma\right) {B}^{m}_{i-1,j-\epsilon} \left(\gamma\right) \label{eq:comamiddle} \\
    &\quad\quad\quad + {B}^m_{i,j-1}\left(\gamma\right) {A}^m_{i-1,j-\epsilon}\left(\gamma\right) \nonumber
\end{alignat}
where $\epsilon+1 \leq j \leq T$ and $A^m_{0,j}=0, B^m_{0,j}=1$ for all $m, j$.
The first term of RHS is the probability that the $m$-th head selects $h_j$ when the head selects a later frame than $j-\epsilon$ in the previous step. 
The second term represents the probability that the $m$-th head chooses $h_j$ when the head does not select any frames between the last selected frame $h_{j-\epsilon}$ and the frame right before the right bound $h_{j-1}$. Thus, equation (\ref{eq:comamiddle}) means the probability for the $m$-th head to choose $h_j$ when predicting $y_i$ so that $\gamma_{i, j} = C^m_{i, j}(\gamma)$.
However, instead of getting $\gamma$ autoregressively, we can replace it with  $\hat{\gamma} = C(\alpha)$  where $\alpha$ is computed by equation (\ref{eq:alpha}) in parallel as below.
\[
 \hat{\gamma}^m_{i,j} = \begin{dcases*}
        {A}^m_{i,j}\left(\alpha\right)  & if $1 \leq j \leq \epsilon$\\
        {C}^m_{i,j}\left(\alpha\right) & if $\epsilon+1 \leq j \leq T$\\
        {B}^m_{i,T}\left(\alpha\right) {B}^m_{i-1,T-\epsilon}\left(\alpha\right) & if $j=T+1$.
        \end{dcases*}
\]
However, this modification has the limitation since word pieces or characters have various length, so constraints on length might be harmful.
Alternatively, we attend the constraints on the difference among heads as \cite{inaguma2020enhancing}.





We suggest our main method called Mutually-Constrained MMA (MCMMA) that estimates the expected alignment with considering the inter-dependency of other MA heads.
Similarly with the equation (\ref{eq:comamiddle}), we define the function $D$ as
\begin{alignat}{1}
    &{D}^m_{i,j}\left(\delta\right) = {A}^m_{i,j}\left(\delta\right) \prod_{m'\neq m} {B}^{m'}_{i,j-\epsilon} \left(\delta\right) \label{eq:hsdelta} \\
    &\quad + {B}^m_{i,j-1}\left(\delta\right) \left( \prod_{m'\neq m} {B}^{m'}_{i,j-\epsilon-1} \left(\delta\right) - \prod_{m'\neq m} {B}^{m'}_{i,j-\epsilon}\left(\delta\right) \right) \nonumber.
\end{alignat}
The first term is the probability that the $m$-th head selects $h_j$ when the other heads do not choose any frames until reaching $h_{j-\epsilon}$. The second term means the probability that the $m$-th head chooses $h_j$ when at least one of the other heads selects $h_{j-\epsilon}$ and the time limit is over. Thus, the equation (\ref{eq:hsdelta}) means the probability for the $m$-th head to choose $h_j$ when predicting $y_i$. Note that the probability to select $h_j$ is zero if at least one of the other heads have chosen $h_o$ where $o<j-\epsilon$.

To avoid training MMA autoregressively, we replace $\delta=D(\delta)$ with $\hat{\delta}$ computed as
\[
 \hat{\delta}^m_{i,j} = \begin{dcases*}
        {A}^m_{i,j}\left(\alpha\right)  & if $1 \leq j \leq \epsilon$\\
        {D}^m_{i,j}\left(\alpha\right) & if $\epsilon+1 \leq j \leq T$\\
        {B}^m_{i,T}\left(\alpha\right) \prod_{m'\neq m} {B}^{m'}_{i,T-\epsilon}\left(\alpha\right) & if $j=T+1$.
        \end{dcases*}
\]

The overall procedures of MMA and MCMMA at inference are in Fig. \ref{fig:concept}. In the training phase, we formulate a context as a weighted sum of the encoder states with the expected alignments. 

\section{Experiment}
\label{sec:experiment}


\subsection{Model Architecture}
\label{ssec:subhead}

Our architecture follows \cite{inaguma2020enhancing} for fairness. 
The model is Transformer-based \cite{DBLP:conf/nips/VaswaniSPUJGKP17}. The encoder is composed of 3 CNN blocks to process audio signals and 12 layers of multi-head self-attention (SAN) with the dimension $d_{model}$ of queries, keys and values, and $H$ heads. Each CNN block comprises 2D-CNN followed by max-pooling with 2-stride and ReLU activation is used after every CNN. 
The decoder has 1D-CNN for positional encoding \cite{mohamed2019transformers} and 6 attention layers. Each of the lower $D_{lm}$ attention layers only has a multi-head SAN followed by a FFN and the upper $(6-D_{lm})$ layers are stacks of SAN, MMA, and FFN. We also adopt chunkwise multihead attention (CA) \cite{inaguma2020enhancing} which provides additional heads for each MA head to consider multiple views of input sequences. $H_{MA}$ and $H_{CA}$ are the number of MA heads per layer and CA heads per MA head, respectively.
The final predictions are obtained after passing through the final linear and softmax layer. Residual connections and layer normalizations are also applied. For the network architecture hyperparameters, We use $(d_{model}, d_{ff}, H, H_{MA}, H_{CA}, D_{lm})$ as $(256,2048,4,4,4, 4)$ and the chunk size of MoChA as 16. We refer the reader to \cite{inaguma2020enhancing} for further details.  


\subsection{Experiment setup}
\label{ssec:subhead}

We experiment on Librispeech 100-hour \cite{DBLP:conf/icassp/PanayotovCPK15} and AISHELL-1 \cite{DBLP:conf/ococosda/BuDNWZ17}.
We implement models on \cite{inaguma2020enhancing}.\footnote{https://github.com/hirofumi0810/neural\_sp}
We utilize the same setup as \cite{inaguma2020enhancing} including extraction for input features and overall experiments for fairness.
We build 10k size of vocabulary by Byte Pair Encoding (BPE). 
Adam optimizer \cite{DBLP:journals/corr/KingmaB14} with Noam learning rate scheduling \cite{DBLP:conf/nips/VaswaniSPUJGKP17}. 
We also adopt the chunk-hopping mechanism \cite{DBLP:conf/icassp/DongWX19} as (the past size, the current size, the future size) = (64, 128, 64) to make encoder streamable. 
We use a pre-trained language model (LM), which is 4-layer LSTM with 1024 units, for inference where the weight of LM and length penalty is 0.5 and 2, respectively with a beam size of 10.
By following \cite{inaguma2020enhancing}, the objective is the negative log-likelihood and the CTC loss with an interpolation weight $\lambda_{ctc} = 0.3$ and the averaged model over the top-10 of models saved at the end of every epoch for final evaluation.
We utilize SpecAugment \cite{DBLP:conf/interspeech/ParkCZCZCL19} for Librispeech, and speed perturbation \cite{DBLP:conf/interspeech/KoPPK15} for AISHELL-1.
Instead of choosing the right bound, we select the most probable frame between the leftmost frame and the right bound in AISHELL-1 from the training stage.
We utilize 2 CNN blocks for encoder and apply max-pooling with 2-stride after the second CNN block, the fourth, and the eighth layer in AISHELL-1.

\subsection{Relative Latency}
\label{ssec:subhead}

Boundary coverage and streamability \cite{inaguma2020enhancing} is the metric to evaluate whether the model is streamable. 
However, it does not well suit with the MMA mechanism since predicting each output is done when the last head completes the selection.
Instead of the above, we utilize the relative latency (ReL) by averaging the difference of hypothesis boundaries from reference boundaries where boundaries are estimated by the alignment of the latest head at each output step. ReL is defined as
\begin{align}
    &\text{ReL}\left(b^{\text{hyp}}, b^{\text{ref}}\right) = \frac{1}{L_{min}} \sum_{i=1}^{L_{min}}\left(b_i^{hyp} - b_i^{ref} \right) \nonumber
\end{align}
where $L_{min}$ is the minimum output length of predictions of the hypothesis and the reference, and $b_i^{hyp}$ and $b_i^{ref}$ are the i-th boundary of the hypothesis and the reference, respectively.

We note that ReL is the natural extension of the existing latency metric. \cite{DBLP:conf/icassp/InagumaGLLG20} provides the utterance-level latency which is the same with ReL when replacing the boundaries produced by the reference model with the gold boundaries in the definition of relative latency. However, acquiring the gold boundaries is complicated, so we utilize the boundaries of MMA without HSD as the reference boundaries.


\begin{table}[t]
\begin{center}
\caption{The comparison of ASR performance on Librispeech and AISHELL-1. $\epsilon=8$ is used in the inference. We do not report the results of MMA with HSD in Librispeech since its WER is too high to compare with the others. * represents the performance that we reproduce. }
\begin{tabular}{|c|l|cc|c|l|}
\hline
\multicolumn{2}{|c|}{\multirow{3}{*}{Model}}                                              & \multicolumn{2}{c|}{\%WER}         & \multicolumn{2}{c|}{\%CER} \\ \cline{3-6} 
\multicolumn{2}{|c|}{} &
  \multicolumn{2}{l|}{\begin{tabular}[c]{@{}l@{}}Librispeech\\ 100-hour\end{tabular}} &
  \multicolumn{2}{l|}{\multirow{2}{*}{\begin{tabular}[c]{@{}l@{}}AISH\\ ELL-1\end{tabular}}} \\ \cline{3-4}
\multicolumn{2}{|c|}{}                                                                    & \multicolumn{1}{c|}{clean} & other & \multicolumn{2}{l|}{}      \\ \hline
\begin{tabular}[c]{@{}c@{}}Off-\\ line\end{tabular} &
  \begin{tabular}[c]{@{}l@{}}Transformer* \\ Transformer \cite{DBLP:conf/interspeech/LuscherBIKMZSN19} \end{tabular} &
  \begin{tabular}[c]{@{}c@{}}11.8 \\ 14.7\end{tabular} &
  \begin{tabular}[c]{@{}c@{}}24.2 \\ 38.5\end{tabular} &
  \multicolumn{2}{c|}{\begin{tabular}[c]{@{}c@{}}5.9\\ -\end{tabular}} \\ \hline
\multirow{5}{*}{\begin{tabular}[c]{@{}c@{}}On-\\ line\end{tabular}} & CBPENC+SBDEC \cite{tsunoo2020streaming}      & -                          & -     & \multicolumn{2}{c|}{7.3}   \\
                                                                    & MMA w/o HSD*          & 12.8                       & 37.0  & \multicolumn{2}{c|}{\textbf{6.2}}   \\
                                                                    & MMA w/ HSD*          & -                       & -  & \multicolumn{2}{c|}{6.3}   \\
                                                                    & MMA+HeadDrop* & 9.0                       & 27.0  & \multicolumn{2}{c|}{6.3}   \\ \cline{2-6} 
                                                                    & MCMMA               & \textbf{8.6}                       & \textbf{24.8}  & \multicolumn{2}{c|}{\textbf{6.2}}   \\ \hline
\end{tabular}
\label{table:mainresult}
\end{center}
\end{table}

\section{Results}
\label{sec:result}

\subsection{Online ASR Results}
\label{ssec:subhead}

We present the results of our approach with baselines in table \ref{table:mainresult}.
We train our model with $\epsilon=10$ and $\epsilon=12$ on Librispeech and AISHELL-1, respectively and evaluate it with $\epsilon=8$ to make the setting same with \cite{inaguma2020enhancing}.
Our model shows better performance than the baselines including HeadDrop \cite{inaguma2020enhancing}.
Especially, we reduce 2.2\% of WER than HeadDrop \cite{inaguma2020enhancing} on test-other in Librispeech.
These results show that training alignments together with other heads' selection time improves the performance.

One very interesting and unexpected point we observed in table \ref{table:mainresult} is that the WER of Transformer is higher than online models (except for MMA) in test-clean experiments. We conjecture that online attention mechanisms are beneficial to exploit locality since they strongly force models to attend small chunks of an input sequence from the training phase.

\subsection{Trade-off between Performance and Latency}
\label{ssec:subhead}

\begin{figure}[t]

\begin{minipage}[b]{1.0\linewidth}
  \centering
  \centerline{\includegraphics[width=1.0\linewidth]{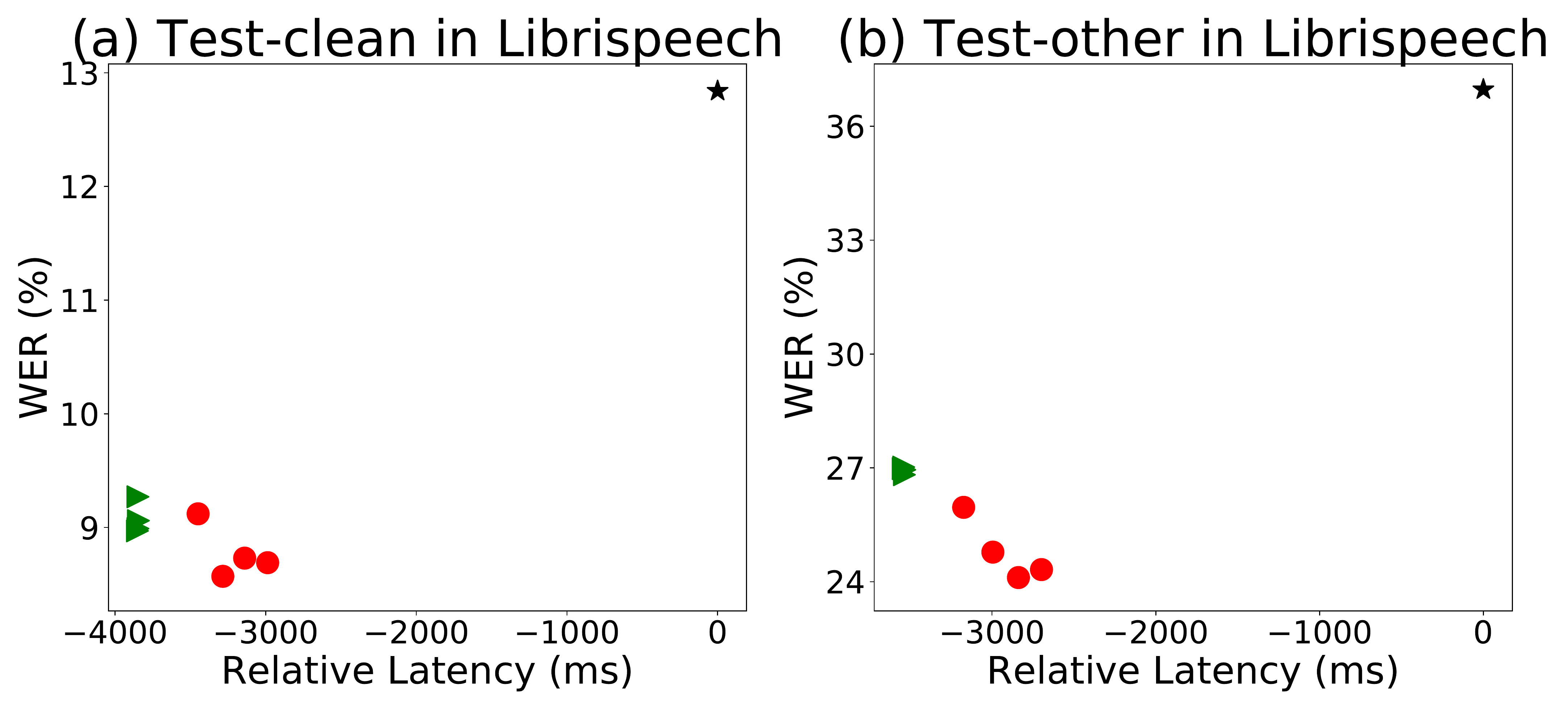}}
\end{minipage}
\hfill
\begin{minipage}[b]{1\linewidth}
  \centering
  \centerline{\includegraphics[width=1\linewidth]{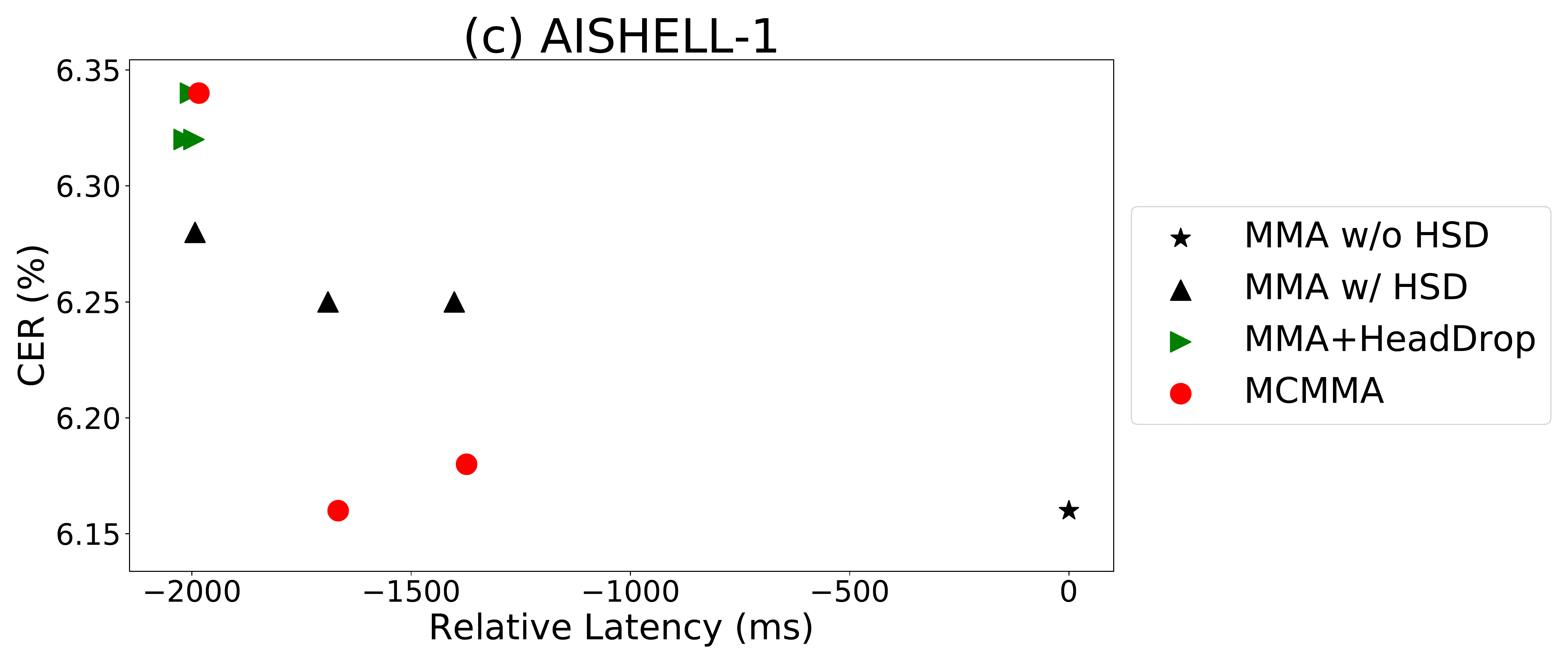}}
\end{minipage}
\caption{The above plots show trade-off between quality and relative latency. (a), (b) are of Librispeech, and (c) is of AISHELL-1.}
\label{fig:tradeoff}
\end{figure}

We provide trade-off graphs between quality and relative latency in fig \ref{fig:tradeoff} through adjusting $\epsilon\in\{6, 8, 10, 12\}$, and $\epsilon\in\{4, 8, 12\}$ in inference time for Librispeech, and AISHELL-1, respectively.
To calculate relative latency with time units, we multiply frame-level relative latency by 80ms since the reducing factor of frames is 8 and the shifting size is 10ms.
Our model outperforms baselines and is still faster than MMA without HSD even though there are small increases in relative latency compared to HeadDrop except for the case with extremely small text $\epsilon$.
The performance degradation with small $\epsilon$ occurs since accessible input information is very limited and training models with small $\epsilon$ restricts head diversity severely.
Thus, this result suggests that the practitioners should avoid choosing small $\epsilon$.

\section{Conclusion}
\label{sec:page}

We suggest the method to learn alignments with considering other heads' alignments by modifying expected alignments for all the heads of each layer to select an input frame within a fixed size window.
Our approach improves performance with only a small increase in latency by regularizing the intra-layer difference of boundaries effectively from the training phase.



\bibliographystyle{IEEEbib}
\bibliography{refs}

\end{document}